\documentclass[runningheads]{llncs}

\PassOptionsToPackage{table}{xcolor}
\usepackage{eccv}

\usepackage{eccvabbrv}

\usepackage{graphicx}
\usepackage{booktabs}
\usepackage{amssymb}
\usepackage{tabularx}
\usepackage{xcolor}
\usepackage{colortbl}

\usepackage[accsupp]{axessibility}  % Improves PDF readability for those with disabilities.

\usepackage{hyperref}

\usepackage{orcidlink}
\usepackage{marvosym}

\definecolor{bestcolor}{RGB}{255,200,200}
\definecolor{secondcolor}{RGB}{255,225,180}
\definecolor{thirdcolor}{RGB}{255,255,180}
\newcommand{\best}[1]{\cellcolor{bestcolor}{#1}}
\newcommand{\second}[1]{\cellcolor{secondcolor}{#1}}
\newcommand{\third}[1]{\cellcolor{thirdcolor}{#1}}

\begin{document}

\title{Habitat-GS: A High-Fidelity Navigation Simulator with Dynamic Gaussian Splatting} 

\titlerunning{Habitat-GS}

\author{Ziyuan Xia\inst{1}\orcidlink{0009-0004-8894-6390} \and
Jingyi Xu\inst{1}\orcidlink{0009-0005-3602-2222} \and
Chong Cui\inst{1}\orcidlink{0009-0007-7130-3745} \and
Yuanhong Yu\inst{1}\orcidlink{0009-0007-7314-2420} \and
Jiazhao Zhang\inst{2}\orcidlink{0000-0001-9459-293X} \and
Qingsong Yan\inst{3}\orcidlink{0000-0002-7095-5844} \and
Tao Ni\inst{4}\orcidlink{0000-0002-8676-6546} \and
Junbo Chen\inst{4}\orcidlink{0009-0002-2640-4137} \and
Xiaowei Zhou\inst{1}\orcidlink{0000-0003-1926-5597} \and
Hujun Bao\inst{1}\orcidlink{0000-0002-2662-0334} \and
Ruizhen Hu\inst{5}\orcidlink{0000-0002-6798-0336} \and
Sida Peng\inst{1}\textsuperscript{\Letter}\orcidlink{0000-0001-6546-4525}}

\authorrunning{Z. Xia et al.}

\institute{Zhejiang University, Hangzhou, China \and
Peking University, Beijing, China \and
XGRIDS, Shenzhen, China \and
UDeer AI, Hangzhou, China \and
Shenzhen University, Shenzhen, China}

\maketitle
{\renewcommand{\thefootnote}{\Letter}\footnotetext{Corresponding author. Email: \email{pengsida@zju.edu.cn}}}

\begin{abstract}
  Training embodied AI agents depends critically on the visual fidelity of simulation environments and the ability to model dynamic humans.
  Current simulators predominantly rely on mesh-based rasterization, for which photorealistic assets are costly to author at scale, and their support for dynamic human avatars is largely constrained to mesh representations, hindering agent generalization to human-populated real-world scenarios.
  We present Habitat-GS, a navigation-centric embodied AI simulator extended from Habitat-Sim that integrates 3D Gaussian Splatting scene rendering and drivable gaussian avatars while maintaining full compatibility with the Habitat ecosystem.
  Our system implements a 3DGS renderer for real-time photorealistic rendering and supports scalable 3DGS asset import from diverse sources.
  For dynamic human modeling, we introduce a gaussian avatar module that enables each avatar to simultaneously serve as a photorealistic visual entity and an effective navigation obstacle, allowing agents to learn human-aware behaviors in realistic settings.
  Experiments on point-goal navigation demonstrate that agents trained on 3DGS scenes achieve stronger cross-domain generalization.
  Evaluations on avatar-aware navigation further confirm that gaussian avatars enable effective human-aware navigation, while performance benchmarks validate the system's scalability.
  Code is available at \url{https://github.com/zju3dv/habitat-gs}.
\end{abstract}

\section{Introduction}

Consider a household service robot tasked with navigating to a target location in a living room where people walk, sit, and move about freely.
To navigate safely and effectively, the robot must perceive its surroundings with sufficient visual fidelity to recognize scenes and objects, and must detect and respond to dynamic human occupants for socially compliant navigation.
The prevailing paradigm in Embodied AI is to train agents at scale in simulation and transfer the learned policies to reality~\cite{anderson2018evaluation,batra2020objectnav,wijmans2019dd}.
This paradigm places two critical demands on the simulator:
the visual realism of rendered sensor observations directly governs the effectiveness of Sim-to-Real policy transfer, and the ability to populate scenes with realistic dynamic humans determines whether agents can learn to navigate safely among people.

\begin{figure*}[t]
    \centering
    \includegraphics[width=\textwidth]{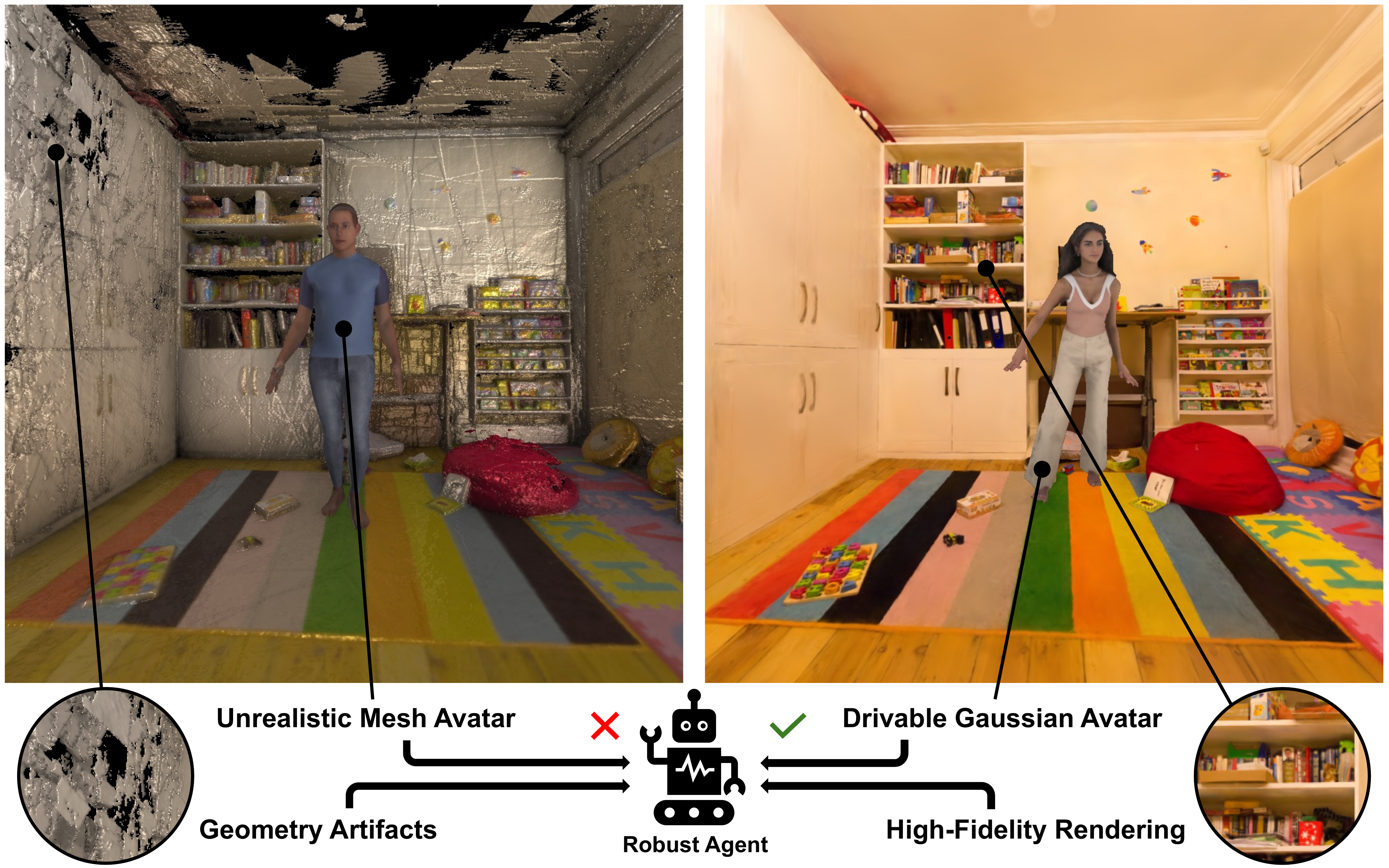}
    \caption[]{\textbf{Habitat-GS} is a navigation-centric embodied simulation platform with 3DGS and dynamic gaussian avatars. Compared to traditional mesh-based simulators (left), our 3DGS-based simulator (right) preserves high-frequency visual details and view-dependent effects, while gaussian avatars provide realistic and dynamic human presence for human-aware navigation scenarios.}
    \label{fig:teaser}
\end{figure*}

The leading open-source Embodied AI simulators, including Habitat-Sim~\cite{habitat1,habitat2,habitat3}, iGibson~\cite{igibson}, AI2-THOR~\cite{ai2thor}, ThreeDWorld~\cite{threedworld}, and SAPIEN~\cite{sapien}, rely on mesh-based rasterization for visual rendering.
While computationally efficient and suited for basic geometric reasoning, the mesh assets used in these simulators are rarely photorealistic, as authoring high-fidelity textured meshes is labor-intensive.
The resulting visual domain gap between simulation and reality, commonly termed the Sim-to-Real gap, is an important factor impacting how well learned navigation policies transfer to physical platforms~\cite{kadian2020sim2real}.
Beyond rendering quality, native support for dynamic high-fidelity human avatars remains limited across these platforms.
Habitat 3.0~\cite{habitat3} introduced mesh-based humanoid avatars for studying social human-robot tasks, yet the visual quality of mesh avatars remains constrained.
Most other simulators provide little or no avatar support, precluding research on navigation in human-populated environments.
Finally, constructing high-quality textured mesh assets from real-world data entails labor-intensive 3D scanning, artist cleanup, and semantic annotation~\cite{replica}, resulting in a scalability bottleneck for expanding the diversity of training environments.

Industrial-grade platforms such as NVIDIA Isaac Sim~\cite{isaac_sim}, built on the Omniverse rendering stack, have recently incorporated 3D Gaussian Splatting (3DGS)~\cite{kerbl3dgs} for enhanced visual fidelity.
However, their closed-source rendering backends impede the deep customization often required for research purposes.
Moreover, these platforms depend on RT Core hardware available only in RTX-series GPUs, limiting their utility on datacenter-class accelerators such as A100 and H100.

We present \textbf{Habitat-GS}, a navigation-centric embodied AI simulator upgraded from Habitat-Sim that integrates 3DGS scene rendering and drivable gaussian avatars while maintaining full compatibility with the Habitat ecosystem, including Habitat-Lab tasks, training, and evaluation APIs.
Habitat-GS addresses the limitations described above along three complementary axes: a 3DGS renderer for real-time photorealistic sensing within the Habitat pipeline, a drivable gaussian avatar module that serves simultaneously as a photorealistic human and an effective navigation obstacle, and scalable import of 3DGS assets from self-reconstructions, public 3DGS datasets~\cite{InteriorGS2025,overmaps}, and generative pipelines such as Marble~\cite{marble}, substantially lowering the barrier to obtaining high-quality scene assets. We validate Habitat-GS through realism studies, performance benchmarks and multiple navigation paradigms.

In summary, our contributions are as follows:
\begin{itemize}
    \item A \textbf{high-fidelity 3DGS-based embodied AI simulation platform} extended from Habitat-Sim for navigation tasks, enabling real-time photorealistic rendering and supporting scalable import from diverse 3DGS asset sources. The platform is fully open-source and maintains complete compatibility with the Habitat ecosystem.
    \item A \textbf{real-time drivable gaussian avatar module} that integrates photorealistic gaussian avatar rendering with collision properties, supporting natural motion and navigation-level obstacle avoidance for human-aware navigation tasks.
    \item \textbf{Comprehensive experimental validation} demonstrates, via VLM-based assessment and PointNav experiments, that 3DGS rendering improves agent cross-domain generalization through a mixed-domain training strategy, and that gaussian avatars enhance agent capability in human-populated scenarios, accompanied by performance benchmarks confirming system scalability.
\end{itemize}

\section{Related Work}

\subsection{Embodied AI Simulators}

Embodied AI research relies on simulation environments for large-scale parallel training and subsequent Sim-to-Real policy transfer~\cite{habitat1,wijmans2019dd}. We compare existing platforms along four dimensions, namely \emph{render asset type}, \emph{humanoid avatar support}, \emph{platform openness}, and \emph{hardware requirements}, in \cref{tab:sim_comparison}.

The predominant embodied AI simulators, including Habitat-Sim~\cite{habitat1,habitat2,habitat3}, iGibson~\cite{igibson,igibson2}, AI2-THOR~\cite{ai2thor}, ThreeDWorld~\cite{threedworld}, and SAPIEN~\cite{sapien}, all employ mesh-based rasterization for scene rendering.
Driving simulators such as CARLA~\cite{carla} confront the same rendering-fidelity challenge, and are increasingly augmented with neural rendering.
Regarding humanoid avatars, while some platforms now provide some form of mesh-based representation, such as URDF-driven articulated bodies~\cite{ai2thor,sapien}, rigid-body tracks~\cite{igibson2}, and Unity Replicants~\cite{threedworld}, these uniformly lack the visual fidelity needed for training agents to perceive fine-grained human appearance and motion cues.
Habitat 3.0~\cite{habitat3} offers the most capable open-source avatar system with deformable SMPL-X mesh avatars for human-robot social tasks, yet mesh rendering inherently constrains visual quality.

Habitat-GS inherits the Habitat ecosystem's high-performance infrastructure and training APIs while upgrading the rendering pipeline from mesh to 3DGS and natively integrating drivable gaussian avatars, uniquely combining photorealistic rendering, dynamic high-fidelity humanoids, and a mature open-source research ecosystem.

\begin{table}[t]
\centering
\caption{\textbf{Comparison of Embodied AI simulation platforms.} ``GPU Req.'' indicates hardware requirements beyond standard CUDA-capable GPUs. ``No Special Req.'' means standard GPUs are sufficient. Habitat-GS is the only platform combining 3DGS rendering with drivable gaussian avatars while remaining fully open-source and deployable on standard datacenter accelerators.}
\label{tab:sim_comparison}
\resizebox{\linewidth}{!}{%
\setlength{\tabcolsep}{2pt}%
\begin{tabular}{@{}lcccc@{}}
\toprule
\textbf{Platform} & \textbf{Render Asset} & \textbf{Humanoid Avatar} & \textbf{Open-Source} & \textbf{GPU Req.} \\
\midrule
Habitat~\cite{habitat1,habitat2,habitat3} & Mesh & Mesh (SMPL-X) & $\checkmark$ & No Special Req. \\
iGibson~\cite{igibson,igibson2} & Mesh & Mesh (Rigid-body) & $\checkmark$ & No Special Req. \\
AI2-THOR~\cite{ai2thor} & Mesh & Mesh (URDF) & $\checkmark$ & No Special Req. \\
ThreeDWorld~\cite{threedworld} & Mesh & Mesh (Replicants) & $\checkmark$ & No Special Req. \\
SAPIEN~\cite{sapien} & Mesh & Mesh (URDF) & $\checkmark$ & No Special Req. \\
Isaac Sim~\cite{isaac_sim} & Mesh+3DGS & Mesh (Sim-Ready) & Partial & Need RT Cores \\
\midrule
\textbf{Habitat-GS (Ours)} & \textbf{Mesh+3DGS} & \textbf{Mesh+GS (SMPL-X)} & $\checkmark$ & No Special Req. \\
\bottomrule
\end{tabular}%
}
\end{table}

\subsection{Neural Rendering}

Neural Radiance Fields (NeRF)~\cite{nerf} demonstrated that implicit neural scene representations can synthesize photorealistic novel views from multi-view images. Subsequent works significantly improved training speed and rendering quality~\cite{instantngp,mipnerf360}. However, NeRF's volume rendering paradigm requires per-pixel ray marching, yielding frame rates far below the real-time requirements of embodied AI sensor pipelines. Moreover, its implicit representation is difficult to integrate with the explicit rendering backends employed by existing simulators.

3D Gaussian Splatting (3DGS)~\cite{kerbl3dgs} represents scenes as collections of explicit anisotropic 3D Gaussians and renders them via differentiable tile-based rasterization, achieving real-time frame rates while maintaining rendering quality comparable to NeRF. The explicit nature of 3DGS lends itself naturally to integration with traditional graphics pipelines, for example via CUDA-OpenGL interop, and facilitates spatial editing, asset composition, and dynamic deformation. Follow-up works have extended 3DGS along multiple directions, including anti-aliasing~\cite{mipsplatting} and dynamic scene modeling~\cite{dynamic3dgaussians}, further broadening its applicability.

\subsection{Gaussian Avatars}

Parametric body models such as SMPL~\cite{smpl} and SMPL-X~\cite{smplx} provide differentiable mappings from pose parameters to human body meshes, serving as the standard infrastructure for avatar animation. Building on this foundation, a growing number of work combines 3DGS with parametric body models to create high-fidelity, drivable human avatars. AnimatableGaussians~\cite{animatablegaussians} maps 2D Gaussians into UV space for pose-driven rendering. GaussianAvatar~\cite{gaussianavatar} deforms canonical-space gaussians to posed configurations via Linear Blend Skinning (LBS). GART~\cite{gart} introduces gaussian articulated templates with differentiable deformation. HumanGaussian~\cite{humangaussian} leverages Score Distillation Sampling~\cite{dreamfusion} for text-driven gaussian avatar generation. The shared paradigm across these methods defines gaussian attributes in a canonical space and transforms them to target poses via skinning. This approach achieves a favorable balance between visual quality and rendering speed that surpasses mesh-based avatars with limited geometric fidelity and NeRF-based avatars with insufficient rendering speed.

\section{Habitat-GS Simulation Environment}\label{sec:method}

Habitat-GS extends Habitat-Sim with two key upgrades: extending mesh-based scene rendering with 3D Gaussian Splatting (\cref{sec:rendering}), and implementing a gaussian avatar module to drive and render avatars (\cref{sec:avatar}).
A central design principle underlying both modules is \emph{visual--navigation decoupling}: 3DGS handles all visual rendering and traditional NavMeshes continue to govern navigation, which sidesteps the lack of explicit surface geometry inherent in the 3DGS representation.
Below we discuss the overall system architecture and the detailed design of each module.

\begin{figure*}[t]
    \centering
    \includegraphics[width=\textwidth]{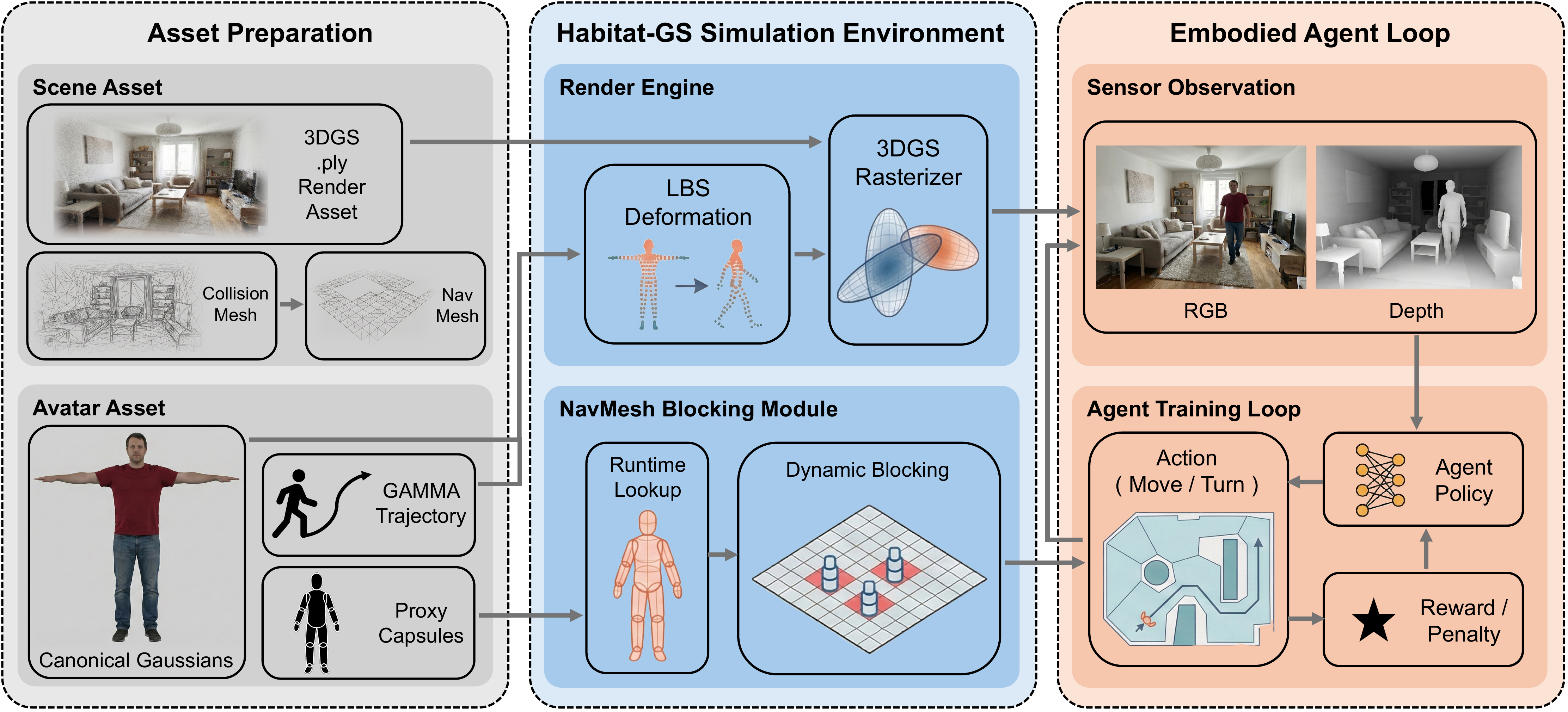}
    \caption{\textbf{System overview of Habitat-GS.} From left to right: \emph{Asset Preparation}, where 3DGS scene assets and gaussian avatar assets are prepared respectively. \emph{Habitat-GS Simulation Environment}, where the render engine performs 3DGS rasterization for scene gaussians and LBS deformation followed by rasterization for avatar gaussians, producing RGB-D observations. The NavMesh blocking module retrieves pre-computed proxy capsules at runtime and injects them into the NavMesh for obstacle blocking. \emph{Embodied Agent Loop}, where the agent policy consumes sensor observations, executes actions, and receives rewards shaped by both navigation success and avatar collision.}
    \label{fig:pipeline}
\end{figure*}

\subsection{System Overview}\label{sec:overview}

\cref{fig:pipeline} illustrates the end-to-end data flow in three stages:

\paragraph{Asset Preparation.}
The scene pipeline prepares 3DGS assets in standard PLY format, together with the conventional NavMesh required by Habitat-Sim's navigation logic.
The avatar pipeline produces three artifacts per identity: \textit{first}, canonical gaussian attributes, specifically positions, SH coefficients, opacities, scales, rotations, and LBS weights, exported from a trained gaussian avatar model. \textit{Second}, GAMMA-generated~\cite{gamma} motion trajectories converted to per-frame SMPL-X joint transformation matrices. \textit{Third}, proxy capsules that approximate the avatar's per-frame collision geometry.

\paragraph{Habitat-GS Simulation Environment.}
At each frame, the system operates two parallel modules.
The \emph{render engine} dispatches scene gaussians to the CUDA rasterizer and simultaneously deforms avatar gaussians to the current pose via CUDA Linear Blend Skinning before rasterizing them with an independent renderer instance. Both color and depth outputs are transferred to OpenGL textures through CUDA--OpenGL interop.
The \emph{navigation module} retrieves the pre-computed proxy capsules for current frame and injects them into the NavMesh as dynamic obstacles, enabling the path planner to produce collision-aware steps around moving avatars.

\paragraph{Embodied Agent Loop.}
The composited RGB-D observations serve as sensor inputs to the agent policy, which outputs discrete actions including move forward, turn left/right, and stop.
The augmented NavMesh ensures that actions violating avatar collision boundaries are clipped, while two avatar state query APIs expose clearance distances and collision flags for reward shaping and metric computation in downstream tasks (\cref{sec:api}).

\subsection{3DGS Scene Rendering Integration}\label{sec:rendering}

\begin{figure}[t]
    \centering
    \includegraphics[width=\linewidth]{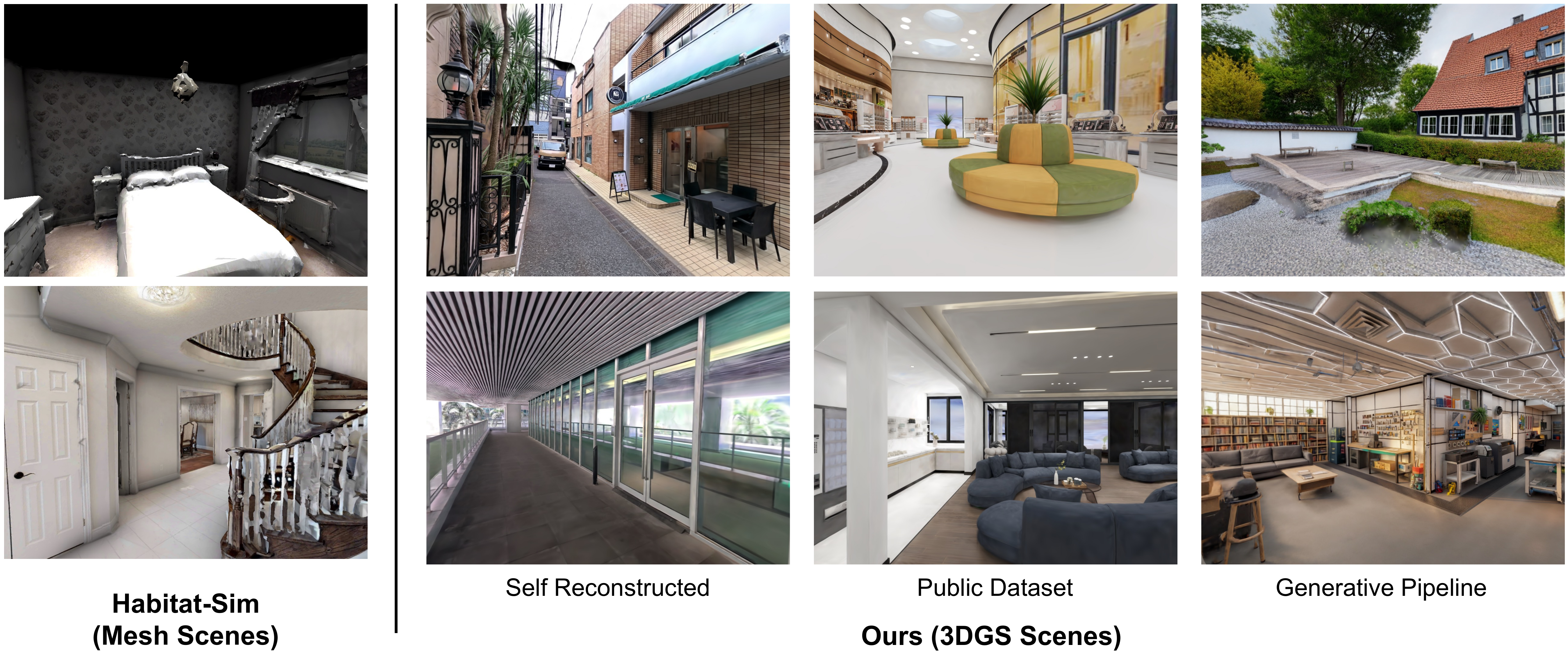}
    \caption{\textbf{Visual comparison of scene rendering.} Mesh-based rendering (left) vs.\ our 3DGS rendering (right). Our simulator is based on 3DGS, which preserves high-frequency details and supports diverse sources of rendering assets.}
    \label{fig:scene_comparison}
\end{figure}

The core technical challenge in integrating 3DGS into Habitat-Sim lies in bridging two heterogeneous rendering backends: the Habitat sensor pipeline is built on OpenGL, whereas high-performance 3DGS rasterization relies on CUDA tile-based rendering~\cite{kerbl3dgs}.
We address this with a zero-copy CUDA--OpenGL interoperability mechanism that keeps all rendering data on the GPU. The CUDA rasterizer writes color and depth directly into pre-registered OpenGL textures, eliminating CPU-side transfer and enabling real-time photorealistic rendering with minimal overhead. As shown in \cref{fig:scene_comparison}, this yields markedly higher visual fidelity than mesh-based rendering. We detail this \emph{Map--Render--Unmap} mechanism in the supplementary material.

A Habitat-GS scene may mix 3DGS assets, traditional meshes, and multiple gaussian avatars.
To resolve occlusion among them, a full-screen pass composites the CUDA-produced 3DGS depth with the OpenGL depth buffer. For each pixel the GS pass yields \mbox{$(C_g, D_g)$} and the mesh pass \mbox{$(C_m, D_m)$}, and the shader keeps the nearer fragment, \mbox{$(C,D)=(D_g<D_m)\,?\,(C_g,D_g):(C_m,D_m)$}.
All active avatars' deformed gaussians are rasterized in a single CUDA pass and composited likewise, giving correct depth ordering among scenes, meshes, and an arbitrary number of avatars within one frame.

\subsection{Dynamic Gaussian Avatar Module}\label{sec:avatar}

\begin{figure}[t]
    \centering
    \includegraphics[width=\linewidth]{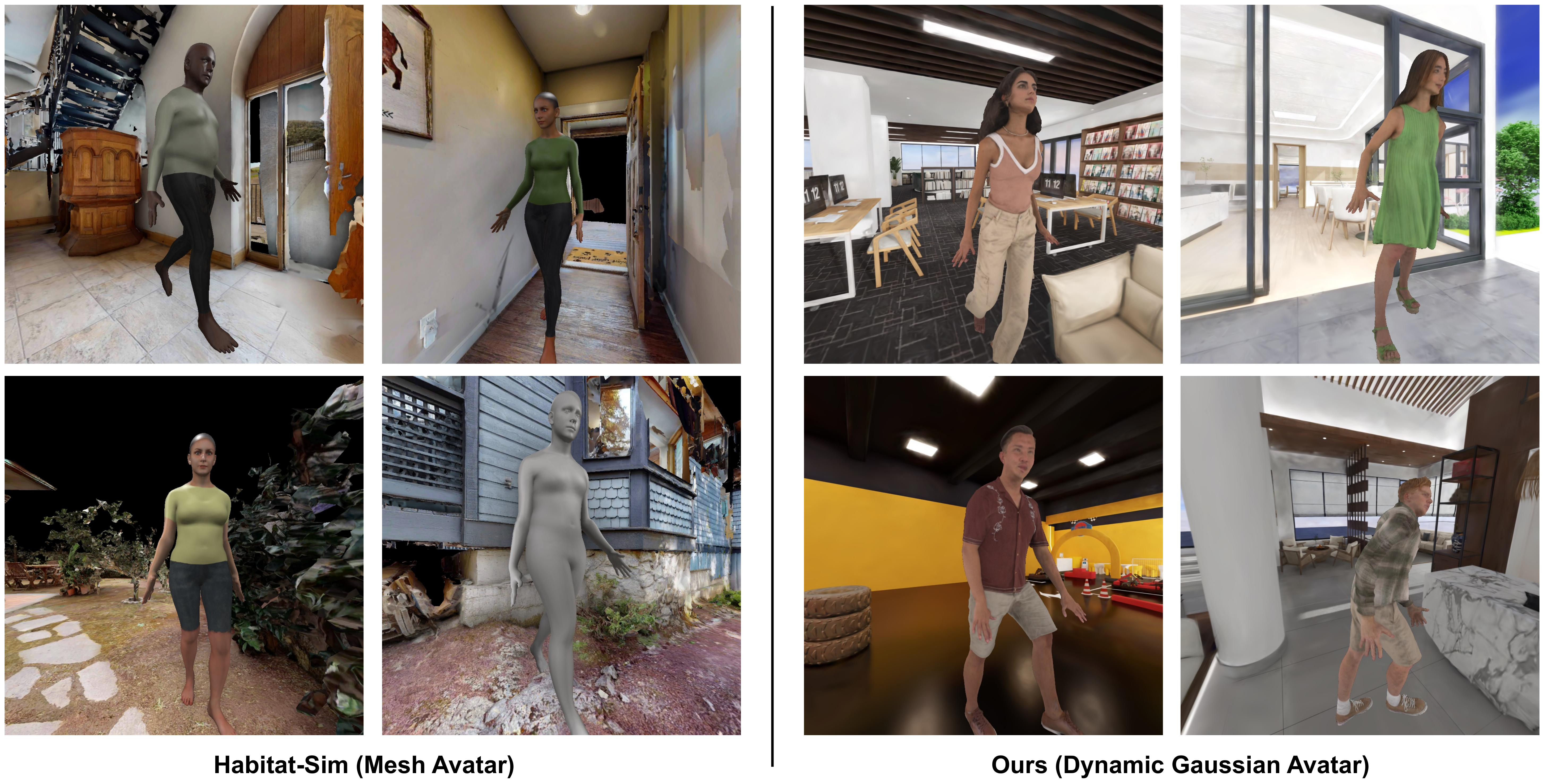}
    \caption{\textbf{Qualitative comparison of mesh avatars and gaussian avatars.} Gaussian avatars exhibit significantly higher visual fidelity, preserving fine-grained details such as clothing wrinkles and hair texture.}
    \label{fig:avatar_comparison}
\end{figure}

Existing gaussian avatar methods~\cite{gaussianavatar,animatablegaussians} focus exclusively on standalone reconstruction and rendering, without coupling to simulator sensor pipelines or navigation mesh systems.
To address this limitation, Habitat-GS incorporates an avatar module that supports the rendering, driving, and collision detection of gaussian avatars, which render with markedly higher visual fidelity than mesh avatars (\cref{fig:avatar_comparison}).
We pre-bake canonical gaussian attributes offline and employ a lightweight CUDA LBS kernel to deform them to arbitrary SMPL-X~\cite{smplx} poses, thereby avoiding costly neural network inference at runtime.
Below we detail how this rendering capability is combined with driving trajectory and NavMesh blocking mechanism.

\subsubsection{GAMMA Trajectory.}

To drive avatars along scene-aware, physically plausible walking paths, we run the GAMMA motion generation model~\cite{gamma} offline on NavMesh shortest paths through sampled waypoints, and convert the resulting poses into per-frame SMPL-X joint transformation matrices via forward kinematics.
At runtime, the avatar module interpolates these matrices at the current simulation time and passes them to the CUDA LBS kernel for deformation.

\subsubsection{Dynamic NavMesh Blocking.}

Since the 3DGS representation lacks a well-defined geometric surface, it is unsuitable for direct collision detection.
We therefore decouple appearance from collision through a \emph{proxy capsule} mechanism. During the offline stage, we derive capsule primitives from the SMPL-X skeletal bones and pre-compute their per-frame world-space positions. At runtime, these are retrieved by temporal interpolation at negligible cost.
On the NavMesh side, to capture the time-varying occupancy of moving avatars, we extend the Habitat-Sim path planner with \emph{dynamic capsule obstacles}. At each step the active avatars' capsules are injected into the planner, which clips any agent motion that would intersect a capsule and thereby guarantees the agent never passes through an avatar's body.
The full avatar kinematics, CUDA LBS, and capsule pre-computation/injection pipelines are detailed in the supplementary material.

\subsection{APIs for Habitat-Lab}\label{sec:api}

Habitat-GS is designed for seamless integration with Habitat-Lab.
During scene initialization, the simulator automatically loads gaussian avatar configurations from the scene description file and instantiates all avatars.
At each simulation step, avatar pose updates are triggered transparently, synchronizing both visual and navigation pipelines.
The sensor outputs, both RGB and depth, produced by 3DGS rendering are format-identical to those of the mesh renderer, allowing existing Habitat-Lab tasks including PointNav~\cite{anderson2018evaluation} to run on 3DGS scenes without modification.

For avatar-aware navigation tasks such as avatar-aware point-goal navigation, two additional query APIs are provided. One returns the distance to the nearest avatar capsule, and the other indicates whether a candidate step is blocked. Together, these provide the information needed for computing rewards and metrics.
Researchers can leverage these APIs to design proximity-based rewards, collision penalties, and tracking metrics within the Habitat-Lab framework.

\section{Experiments}

We evaluate Habitat-GS along three complementary axes:
\textit{First}, we demonstrate that 3DGS scenes are more photorealistic than mesh scenes through VLM, human, and feature-space assessments (\cref{sec:render_quality}), and that training on these higher-quality scenes improves agent cross-domain generalization and Sim-to-Real transfer on PointNav (\cref{sec:static_nav}).
\textit{Second}, we examine whether training with dynamic gaussian avatars equips agents with human-aware navigation capabilities on dynamic point-goal navigation (\cref{sec:dynamic_nav}).
\textit{Third}, we examine whether the system remains efficient under varying scene complexity and avatar counts (\cref{sec:perf}).
We first describe the shared experimental setup.

\subsection{Experimental Setup}\label{sec:exp_setup}

\subsubsection{Datasets.}
For \textit{3DGS scenes}, we combine the InteriorGS dataset~\cite{InteriorGS2025} with additional real-world reconstructed GS scenes in a 4:1 ratio, yielding 120 scenes split into 100 for training and 20 for testing.
For \textit{mesh scenes}, we use the Habitat-Matterport 3D (HM3D) dataset~\cite{hm3d}, similarly split into 100 training and 20 test scenes.
Importantly, the GS and mesh test sets are drawn from disjoint scene collections rather than two representations of the same physical spaces, so that evaluation measures cross-domain generalization.

For \textit{gaussian avatars}, we export canonical gaussians from six trained AnimatableGaussians~\cite{animatablegaussians} identities, with three for training and three held out for testing. Each avatar is driven by GAMMA-generated~\cite{gamma} motion trajectories with pre-computed joint matrices and proxy capsules.

\subsubsection{Evaluation Metrics.}

We adopt standard embodied navigation metrics.
\textbf{Success Rate (SR)} measures the fraction of episodes in which the agent reaches the goal within a distance threshold.
\textbf{Success weighted by Path Length (SPL)}~\cite{anderson2018evaluation} jointly captures success and path efficiency: $\text{SPL} = \text{SR} \times (\ell^{*} / \ell)$, where $\ell^{*}$ is the geodesic shortest-path length and $\ell$ the agent's actual path length.
\textbf{Distance to Goal (DTG)} records the Euclidean distance between the agent and the goal at episode termination. Lower values indicate the agent consistently navigates closer to the target even in unsuccessful episodes.

For avatar-aware tasks, we additionally report:
\textbf{Collision Rate (CR)}, the fraction of collision steps.
\textbf{Personal Space Intrusion (PSI)}, which quantifies the average degree to which the agent enters the 1.0\,m personal-space radius around each avatar. Higher values indicate more frequent and severe intrusions.

\subsection{3DGS Scenes Deliver Higher Quality Rendering}\label{sec:render_quality}

\begin{figure}[t]
    \centering
    \includegraphics[width=\linewidth]{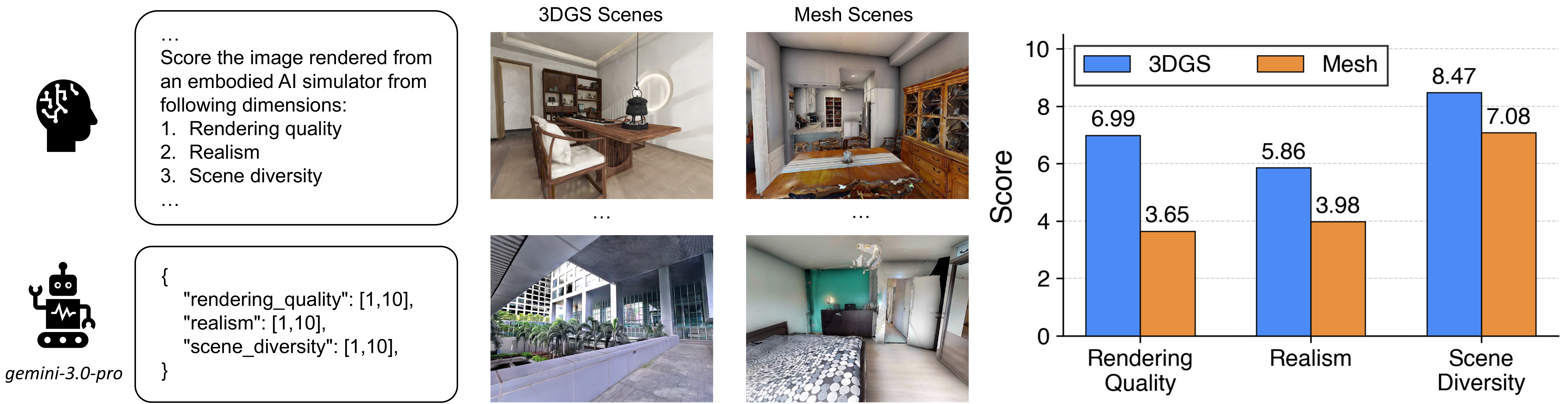}
    \caption{\textbf{VLM scene quality assessment.} Gemini 3.0 Pro evaluates 240 rendered screenshots from each domain on three perceptual dimensions. GS scenes consistently outperform mesh scenes, confirming their superior visual fidelity and diversity.}
    \label{fig:vlm_eval}
\end{figure}

We assess the rendering quality of 3DGS scenes and mesh scenes from three complementary angles: an automated VLM judge, a study with human raters, and a quantitative comparison to real photographs in feature space.

\subsubsection{VLM Scene Quality Assessment.}
To objectively quantify the quality gap between 3DGS and mesh scene rendering, we employ Gemini 3.0 Pro~\cite{gemini2024} as an automated evaluator.
We render 240 screenshots evenly from each renderer, divided into 48 evaluation batches, each containing 5 GS and 5 mesh images with randomized indices to blind the model of rendering source.
The VLM scores each image on a 10-point scale along three dimensions: \emph{rendering quality}, \emph{realism}, and \emph{scene diversity}.
As shown in \cref{fig:vlm_eval}, GS scenes score higher than mesh scenes across all three dimensions in this automated assessment.

\subsubsection{Human Perceptual Study.}
We next corroborate the evaluation with a study using human raters. Each batch presents $3$ GS and $3$ mesh renderings in randomized order, and raters are asked to rank them by perceived realism. Across $88$ participants and $825$ valid batches with $7425$ GS-vs-mesh pairwise comparisons, GS renderings attain a markedly better mean rank and are preferred in $61.9\%$ of pairwise comparisons, which strongly supports the realism advantage of 3DGS rendering.

\begin{table}[t]
\centering
\caption[Realism evaluation of 3DGS vs.\ mesh rendering]{\textbf{Realism evaluation of 3DGS vs.\ mesh rendering.}
\textbf{(a)}~Human ranking study.
\textbf{(b)}~Distance of GS / mesh renderings to a pool of real photographs in CLIP and DINOv2 feature spaces.
\colorbox{bestcolor}{Best} (more realistic / closer to real) is highlighted.}
\label{tab:realism_studies}
\setlength{\fboxsep}{3pt}%
\begin{minipage}[t]{0.34\linewidth}
\centering
{\small\textbf{(a) Human perceptual ranking}}\\[4pt]
\resizebox{\linewidth}{!}{%
\setlength{\tabcolsep}{6pt}%
\begin{tabular}{@{}l cc@{}}
\toprule
 & GS & Mesh \\
\midrule
Mean rank ($\downarrow$)    & \best{3.14}   & 3.86 \\
Pairwise wins ($\uparrow$)  & \best{61.9\%} & 38.1\% \\
\bottomrule
\end{tabular}}
\end{minipage}%
\hspace{0.03\linewidth}%
\begin{minipage}[t]{0.60\linewidth}
\centering
{\small\textbf{(b) Feature-space distance to real photos ($\downarrow$)}}\\[4pt]
\resizebox{\linewidth}{!}{%
\setlength{\tabcolsep}{6pt}%
\begin{tabular}{@{}l cc cc@{}}
\toprule
 & \multicolumn{2}{c}{CLIP} & \multicolumn{2}{c}{DINOv2} \\
\cmidrule(lr){2-3}\cmidrule(lr){4-5}
Real reference \& metric & GS & Mesh & GS & Mesh \\
\midrule
InCrowd-VI~\cite{incrowdvi} / cos  & \best{0.252} & 0.343 & \best{0.583} & 0.729 \\
InCrowd-VI~\cite{incrowdvi} / CMMD & \best{0.131} & 0.186 & \best{0.108} & 0.148 \\
Self-captured / cos                & \best{0.155} & 0.256 & \best{0.470} & 0.772 \\
Self-captured / CMMD               & \best{0.068} & 0.123 & \best{0.052} & 0.104 \\
\bottomrule
\end{tabular}}
\end{minipage}%

\end{table}

\subsubsection{Feature-Space Realism Evaluation.}
As an objective complement, we measure how close each renderer is to \emph{real} photographs in learned feature spaces. Using $240$ GS and $240$ mesh renderings and a pool of $480$ real photos ($240$ InCrowd-VI~\cite{incrowdvi} and $240$ self-captured), we compute centroid cosine distance and CMMD~\cite{cmmd} in CLIP~\cite{clip} and DINOv2~\cite{dinov2} feature spaces, where a smaller distance means closer to real. As shown in \cref{tab:realism_studies}(b), GS renderings are closer to real photographs than mesh renderings on every backbone, metric, and reference pool. Taken together, the VLM, human and feature-space assessments all give consistent evidence that 3DGS scenes deliver more realistic and higher quality rendering than mesh scenes under comparable asset-acquisition effort.

\subsection{3DGS Scenes Train More Generalizable Agents}\label{sec:static_nav}

We then test whether the rendering-quality advantage yields navigation agents that generalize better, at two levels: across simulation domains (Sim-to-Sim) and from simulation to real-world video (Sim-to-Real).

\subsubsection{Sim-to-Sim PointNav Evaluation.}
We train five agent groups under different scene-domain mixtures on the \textbf{PointNav}~\cite{anderson2018evaluation} task, with training budget fixed at $5{\times}10^{7}$ steps:
\textbf{A}:~100 mesh scenes,
\textbf{B}:~100 GS scenes,
\textbf{C}:~80\,M + 20\,G,
\textbf{D}:~50\,M + 50\,G, and
\textbf{E}:~20\,M + 80\,G.
All agents share a unified DD-PPO~\cite{wijmans2019dd} architecture with a ResNet~\cite{resnet} visual encoder and a GRU~\cite{cho2014gru} policy head, receiving $256{\times}256$ RGB and depth observations, with only training scene composition varying.
Each agent is evaluated on both the 20-scene mesh test set and the 20-scene GS test set, so that performance on the unseen domain measures cross-domain generalization.

\begin{table}[t]
\centering
\caption{\textbf{PointNav results and training curves.}
(a)~Results across different training domain mixtures. Agents are trained on varying ratios of mesh (M) and 3DGS (G) scenes and evaluated on disjoint mesh and GS test sets. SR and SPL are in \%; DTG is in meters ($\downarrow$).
\colorbox{bestcolor}{Best} and \colorbox{secondcolor}{Second best} results per column are highlighted.
(b)~Training success rate curve for 100\,Mesh (Config A).
(c)~Training success rate curve for 100\,GS (Config B).}
\label{tab:static_nav}

\begin{minipage}[t]{0.73\linewidth}
\centering
\textbf{(a) Quantitative Results}\\[2pt]
\resizebox{\linewidth}{!}{%
\setlength{\tabcolsep}{2.5pt}%
\begin{tabular}{@{}l ccc ccc@{}}
\toprule
& \multicolumn{3}{c}{\textbf{Mesh Test}} & \multicolumn{3}{c}{\textbf{GS Test}} \\
\cmidrule(lr){2-4} \cmidrule(lr){5-7}
\textbf{Training Config}
    & SR$\uparrow$ & SPL$\uparrow$ & DTG$\downarrow$
    & SR$\uparrow$ & SPL$\uparrow$ & DTG$\downarrow$ \\
\midrule
A: 100\,Mesh
    & 59.00 & 51.23 & \best{5.537}
    & 61.30 & 52.09 & 4.982 \\
B: 100\,GS
    & 53.00 & 43.13 & 6.439
    & 70.70 & 58.49 & 3.550 \\
C: 80\,M + 20\,G
    & \second{60.20} & \best{52.06} & 6.004
    & 73.40 & 64.32 & 3.757 \\
D: 50\,M + 50\,G
    & \best{61.80} & \second{51.34} & 5.938
    & \second{78.10} & \second{67.42} & \second{3.008} \\
E: 20\,M + 80\,G
    & 59.60 & 51.01 & \second{5.901}
    & \best{79.60} & \best{68.38} & \best{2.698} \\
\bottomrule
\end{tabular}%
}
\end{minipage}%
\hfill
\begin{minipage}[t]{0.25\linewidth}
\centering
\textbf{(b) Mesh Training}\\[2pt]
\includegraphics[width=\linewidth]{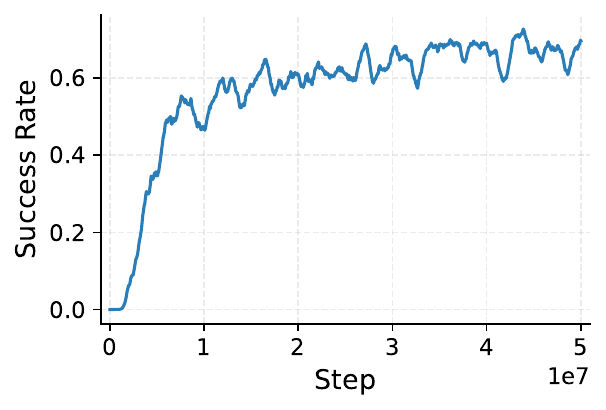}

\textbf{(c) GS Training}\\[2pt]
\includegraphics[width=\linewidth]{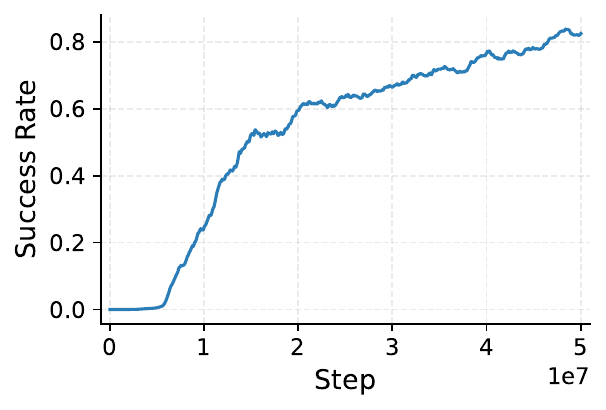}
\end{minipage}
\end{table}

\subsubsection{Sim-to-Real Validation on Real-World Videos.}
To further evaluate the robustness of agents trained on different domains, we run an open-loop PointNav evaluation on real videos: $5$ self-recorded scenes (Record3D / iPhone, $38$ episodes) and $5$ InCrowd-VI~\cite{incrowdvi} scenes (Aria glasses, $25$ episodes). At each frame we compare the trained agent's discrete action to the human's actual next motion and report action-prediction agreement with a $\pm1$-step tolerance (APA$\pm1$). This open-loop protocol probes whether the agent's predicted actions align with real human trajectories, testing the Sim-to-Real transfer capability.

\begin{table}[t]
\centering
\caption[Real-video open-loop PointNav APA-pm-1]{\textbf{Real-video open-loop PointNav Evaluation.}
At each frame we compare the trained agent's discrete action to the human's actual next motion and report APA$\pm1$ (\%, $\uparrow$).
\textbf{N} is the number of episodes per scene.
\colorbox{bestcolor}{Best}, \colorbox{secondcolor}{2nd}, and \colorbox{thirdcolor}{3rd} in each scene column are highlighted.
The \emph{mixed} $>$ \emph{GS} $>$ \emph{mesh} ordering on real video mirrors the cross-domain trend in \cref{tab:static_nav}.}
\label{tab:rw_apa}
\setlength{\fboxsep}{1.5pt}%
\resizebox{0.98\linewidth}{!}{%
\setlength{\tabcolsep}{2pt}%
\renewcommand{\arraystretch}{0.95}%
\begin{tabular}{@{}l|ccccc|c|ccccc|c@{}}
\toprule
& \multicolumn{6}{c|}{\textit{Self-recorded (Record3D, iPhone)}}
& \multicolumn{6}{c}{\textit{InCrowd-VI (Aria glasses)}} \\
\cmidrule(lr){2-7}\cmidrule(lr){8-13}
\textbf{Scene}
    & yard & lab. & gard. & off. & park. & \textbf{\emph{Avg}}
    & Zur. & BIN  & UZH  & TS   & AND  & \textbf{\emph{Avg}} \\
\textbf{N}
    & 9    & 9    & 2     & 9    & 9     & 38
    & 3    & 4    & 8    & 4    & 6    & 25 \\
\midrule
A: 100\,Mesh    & 66.2 & \second{28.6} & \best{78.5} & 61.0 & 61.4 & 55.6 & 33.8 & 11.5 & 8.4 & \second{47.9} & 17.1 & 20.4 \\
B: 100\,GS      & \third{78.1} & \third{25.2} & \best{78.5} & \third{62.6} & \third{61.8} & 58.1 & 44.4 & \best{18.8} & \best{16.9} & 41.1 & \best{24.1} & \third{26.1} \\
C: 80\,M + 20\,G     & 76.1 & 24.5 & 74.0 & 62.1 & \second{69.2} & \third{58.8} & \third{70.3} & \third{13.3} & 9.7 & 41.4 & 17.2 & 24.4 \\
D: 50\,M + 50\,G     & \second{79.3} & \best{29.2} & \second{77.0} & \best{65.7} & 61.7 & \second{59.9} & \second{71.5} & 12.5 & \second{15.3} & \best{60.3} & \third{18.8} & \best{29.6} \\
E: 20\,M + 80\,G     & \best{81.6} & 23.7 & \third{75.0} & \second{64.3} & \best{72.0} & \best{61.2} & \best{73.5} & \second{17.5} & \third{12.4} & \third{45.6} & \second{22.4} & \second{28.3} \\
\bottomrule
\end{tabular}}
\end{table}

\subsubsection{Analysis.}

\cref{tab:static_nav} and \cref{tab:rw_apa} reveal a clear picture of three training strategies:
\paragraph{Mesh-only training converges quickly but generalizes poorly.}
As seen in \cref{tab:static_nav}(b), training on 100 mesh scenes converges rapidly, yet the resulting agent plateaus at a modest success rate around 0.6 and generalizes poorly to the GS test set.
The same ranking holds under Sim-to-Real in \cref{tab:rw_apa}. The mesh-only agent attains the lowest real-video APA$\pm1$ on both sources, confirming that it transfers least well to reality.
Mesh-only training therefore generalizes least effectively to realistic visual domains within this budget.
\paragraph{GS-only training builds stronger capability but converges too slowly.}
\cref{tab:rw_apa} shows that training on 3DGS scenes can substantially improve performance on Sim-to-Real evaluation compared with mesh-only training, proving that the added visual realism aids real-world transfer.
However, because GS scenes are more realistic and diverse, the agent requires considerably more steps to converge.
Within the fixed budget of $5{\times}10^{7}$ steps, the GS-only agent has not fully converged according to \cref{tab:static_nav}(c), and its mesh test performance remains below that of Config~\textbf{A}.
Pure GS training is therefore not cost-effective under practical compute constraints.
\paragraph{Mixed-domain training is the most effective strategy.}
Configurations \textbf{D} (50\,M + 50\,G) and \textbf{E} (20\,M + 80\,G) demonstrate the most effective approach: a small number of mesh scenes first establish foundational navigation competence, after which training on more realistic GS scenes further strengthens the agent's ability to handle diverse and realistic environments.
Config~\textbf{E} achieves the best GS test performance while maintaining mesh test performance on par with the mesh-only baseline.
The same advantage carries over to Sim-to-Real in \cref{tab:rw_apa}: the mixed-domain agents attain the best real-video APA$\pm1$ overall, with Config~\textbf{E} reaching $61.2$ on the self-recorded set and Config~\textbf{D} reaching $29.6$ on InCrowd-VI. This offers convincing evidence that 3DGS training helps narrow the visual Sim-to-Real gap.
These results together indicate that the two data modalities are complementary: mesh scenes provide efficient geometric learning, while GS scenes contribute the visual realism that transfers to more realistic deployment environments.

\subsection{Gaussian Avatars Equip Agents with Human-Aware Navigation}\label{sec:dynamic_nav}

We next evaluate whether gaussian avatars enable agents to navigate safely among dynamic humans.
We extend the standard PointNav task by introducing three gaussian avatars walking around in scenes, and agents are supposed to reach the goal point without colliding with moving avatars.

The agents are first trained on static PointNav for $5{\times}10^{7}$ steps to establish basic navigation competence, and then fine-tuned for an additional $5{\times}10^{6}$ steps on a subset of 20 GS scenes under two configurations:
\textbf{Baseline}:~GS scenes, no avatars; and
\textbf{GS Scene + GS Avatar}:~GS scenes with gaussian avatars.
Both groups are evaluated on two dynamic test sets to assess cross-domain capability: 20 mesh scenes with mesh avatars, and 20 GS scenes with gaussian avatars.
We prioritize \textbf{CR} and \textbf{PSI} in avatar-aware navigation evaluation, for collision is unacceptable in real-world human-aware navigation.

\begin{table}[t]
\centering
\caption{\textbf{Avatar-aware PointNav results.}
\textbf{Top:} Two training configurations are evaluated: a GS scene baseline without avatars, and a GS scene populated with GS avatars.
Agents are evaluated on both mesh and GS dynamic test sets to assess in-domain performance and cross-domain generalization.
SR, SPL and CR are in \%. CR is the fraction of collision steps ($\downarrow$). PSI is the average personal-space intrusion ($\downarrow$).
\textbf{Bottom:} Visual demonstration of our avatar-aware navigation policy. Agent successfully navigates through gaussian avatars without collision or personal space intrusion.
}
\label{tab:dynamic_nav}

\resizebox{\linewidth}{!}{%
\setlength{\tabcolsep}{4pt}%
\begin{tabular}{@{}l cccc cccc@{}}
\toprule
& \multicolumn{4}{c}{\textbf{Mesh Scene+Avatar Test}} & \multicolumn{4}{c}{\textbf{GS Scene+Avatar Test}} \\
\cmidrule(lr){2-5} \cmidrule(lr){6-9}
\textbf{Training Config}
    & SR$\uparrow$ & SPL$\uparrow$ & CR$\downarrow$ & PSI$\downarrow$
    & SR$\uparrow$ & SPL$\uparrow$ & CR$\downarrow$ & PSI$\downarrow$ \\
\midrule
GS Baseline                     & 54.80 & 46.98 & 2.521 & 0.075 & 81.80 & 71.35 & 6.713 & 0.092 \\
GS Scene+GS Avatar         & 58.00 & 46.80 & 2.342 & 0.068 & 80.00 & 65.72 & 4.746 & 0.077 \\
\bottomrule
\end{tabular}%
}

\begin{tabular}{@{}c@{\hspace{4pt}}c@{}}
\includegraphics[width=0.485\linewidth]{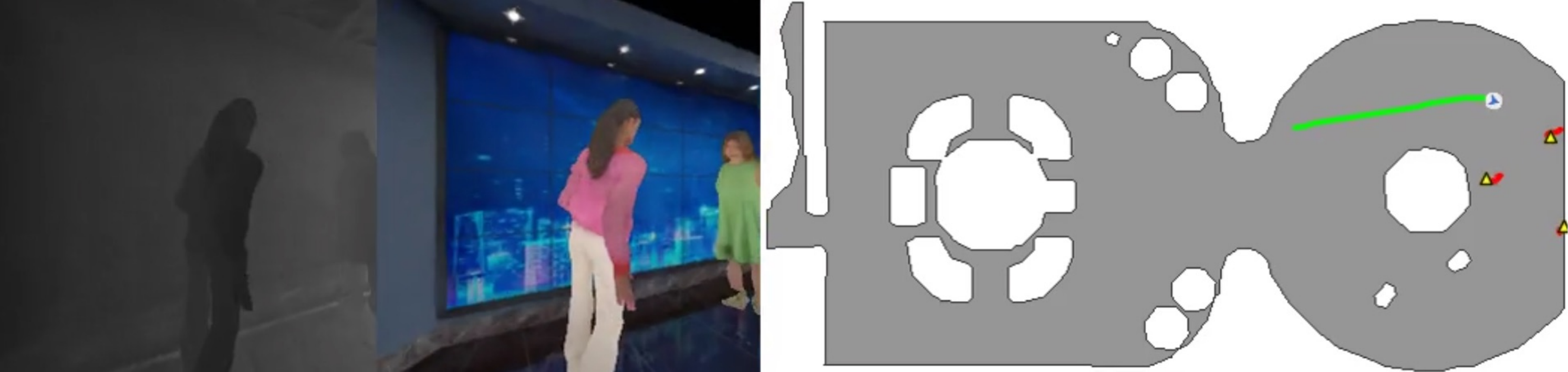} & \includegraphics[width=0.485\linewidth]{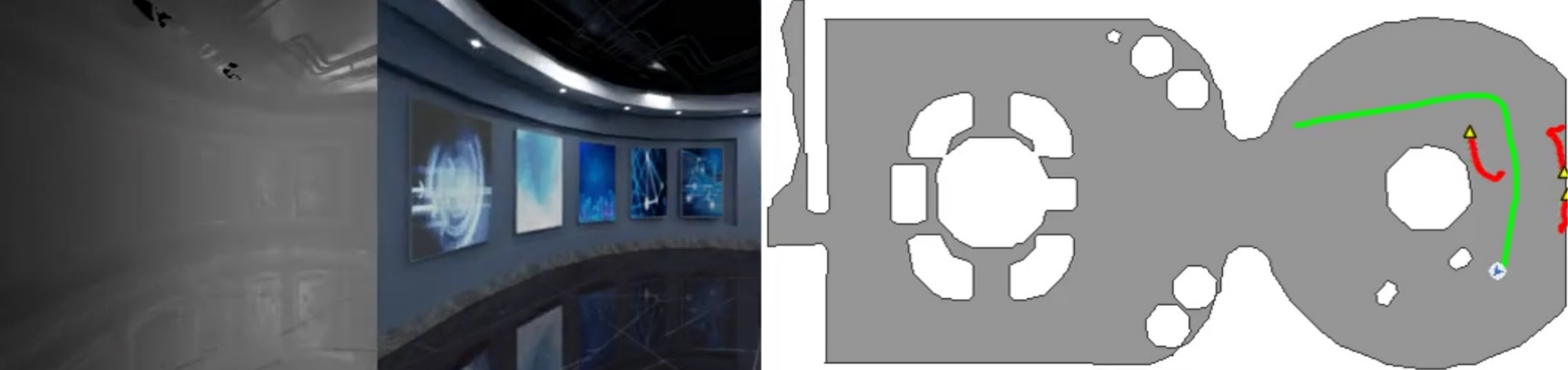} \\
\small{\textbf{(a) Start of episode}} & \small{\textbf{(b) End of episode}}
\end{tabular}
\end{table}

\subsubsection{Analysis.}

\cref{tab:dynamic_nav} demonstrates the clear benefit of training with gaussian avatars for human-aware navigation.
\textit{First}, compared to the avatar-free baseline, training with gaussian avatars effectively lowers the Collision Rate (CR) and Personal Space Intrusion (PSI) on the in-domain GS test set. This human-aware navigation capability is acquired with only a small number of fine-tuning steps ($5{\times}10^{6}$, which is only 10\% of the static pre-training budget), confirming that agents efficiently learn collision avoidance behaviors from the photorealistic humans and realistic motion simulations.
\textit{Second}, the agent fine-tuned with gaussian avatars also achieves lower CR (2.342\% vs. 2.521\%) and PSI (0.068 vs. 0.075) on the mesh test set. This cross-domain improvement suggests that the perceptual cues learned from gaussian avatars, such as recognizing human shape, anticipating walking direction, and estimating personal space, transfer effectively even to lower-fidelity mesh environments.

\subsection{Performance Benchmarks}\label{sec:perf}

To confirm that Habitat-GS's photorealistic rendering does not compromise training speed significantly, we benchmark rendering speed and GPU memory under varying scene scales and avatar counts.
All measurements are conducted on Habitat-Lab's PointNav training on a single NVIDIA RTX 4090 GPU at $256{\times}256$ resolution, matching the typical training configuration.

\begin{table}[t]
\centering
\caption{\textbf{Rendering performance and GPU memory usage.}
Mesh Ref.\ reports Habitat-Sim's native mesh rasterization on a representative HM3D scene as a baseline.
(a)~Varying scene Gaussian count with no avatars.
(b)~Varying avatar count in a medium-scale scene with approximately 2M Gaussians.
Habitat-GS maintains real-time throughput under typical RL training loads with up to 5M Gaussians and $<$5 avatars.}
\label{tab:perf}
\resizebox{\linewidth}{!}{%
\setlength{\tabcolsep}{3pt}%
\begin{tabular}{@{} l c cccccc ccccc @{}}
\toprule
& \multicolumn{1}{c}{\textbf{Mesh}} & \multicolumn{6}{c}{\textbf{(a) Scene Scale (Gaussian Count)}} & \multicolumn{5}{c}{\textbf{(b) Avatar Count}} \\
\cmidrule(lr){2-2} \cmidrule(lr){3-8} \cmidrule(lr){9-13}
& Ref. & 300\,K & 500\,K & 1\,M & 3\,M & 5\,M & 7\,M & 0 & 1 & 2 & 5 & 10 \\
\midrule
FPS$\uparrow$       & 163.8 & 159.2 & 148.8 & 120.9 & 82.60 & 51.46 & 44.52 & 94.16 & 75.14 & 57.70 & 37.74 & 24.67 \\
Frame time (ms)$\downarrow$ & 6.11 & 6.28 & 6.72 & 8.27 & 12.1 & 19.4 & 22.5 & 10.6 & 13.3 & 17.3 & 26.5 & 40.5 \\
Mem (GB)$\downarrow$ & 2.930 & 3.299 & 3.373 & 3.567 & 3.723 & 4.091 & 4.425 & 3.917 & 4.197 & 4.457 & 5.513 & 7.497 \\
\bottomrule
\end{tabular}%
}
\end{table}

\subsubsection{Analysis.}
\cref{tab:perf} demonstrates two key properties of the system.
\textit{First}, under typical RL training loads, Habitat-GS sustains real-time rendering sufficient for large-scale DD-PPO training.
For medium-scale scenes with 1--2 avatars, the system still maintains $>$50 FPS, well above the threshold for efficient parallel training.
\textit{Second}, GPU memory consumption grows approximately linearly with both scene gaussian count and avatar count, providing predictable resource budgeting.
The throughput reduction of 3DGS rendering, relative to mesh-based rasterization, is a favorable trade-off given the improvements in visual fidelity and agent generalization demonstrated in \cref{sec:render_quality} and \cref{sec:static_nav}.

\section{Conclusion}

We present Habitat-GS, a navigation-centric embodied AI simulator that upgrades the visual backbone of Habitat-Sim from mesh rasterization to 3D Gaussian Splatting, and populates environments with photorealistic drivable gaussian avatars.
Habitat-GS is fully open-sourced to provide a high-fidelity, ecosystem-compatible foundation for future embodied AI research.

\noindent\textbf{Limitations.}
A central design choice of Habitat-GS is \emph{visual--navigation decoupling} (\cref{sec:overview}): 3DGS handles all visual rendering while NavMeshes are responsible for navigation logic.
This intentional decoupling sidesteps the absence of explicit surface geometry in the 3DGS representation. However, it also confines physical interaction to navigation-level obstacle avoidance rather than force- or impulse-level contact.
Habitat-GS is therefore best suited for navigation tasks. Extending support to manipulate GS-represented objects would require deeper integration with the physics engine, which we identify as a promising direction for future work.
In addition, avatars in Habitat-GS are currently driven along pre-synthesized trajectories and act as high-priority dynamic obstacles rather than interactive co-agents. Coupling avatars to agent behavior at runtime would require neural inference in the loop and is left to future work.

\section*{Acknowledgements}

This work was partially supported by NSFC (No. U24B20154, No. 62402427), Zhejiang Provincial Natural Science Foundation of China under Grant No. LZ26\-F020003, “Innovation Yongjiang 2035” Key R\&D Program (No. 2025Z061) and Information Technology Center and State Key Lab of CAD\&CG, Zhejiang University.
\bibliographystyle{splncs04}
\bibliography{main}
\end{document}